\documentclass[lettersize,journal]{IEEEtran}
\usepackage{amsmath,amsfonts}
\usepackage{algorithmic}
\usepackage{algorithm}
\usepackage{array}
\usepackage{arydshln}
\usepackage{booktabs}
\usepackage[caption=false,font=normalsize,labelfont=sf,textfont=sf]{subfig}
\usepackage{cases}
\usepackage{color}
\usepackage{float}
\usepackage{makecell}
\usepackage{nicematrix}
\usepackage{multirow}
\usepackage{ragged2e}
\usepackage{subfloat}
\usepackage{stfloats}
\usepackage{textcomp}
\usepackage{url}
\usepackage{verbatim}
\usepackage{graphicx}
\usepackage{cite}

\usepackage[colorlinks,linkcolor=blue]{hyperref}

\def\onedot{. }
 
\def\ie{\emph{i.e}\onedot}

\def\etal{\emph{et al}\onedot}

\begin{document}

\title{Pre-train, Adapt and Detect: Multi-Task Adapter Tuning for Camouflaged Object Detection}

\author{Yinghui Xing*,
        Dexuan Kong*,
        Shizhou Zhang$^{\dag}$,
        Geng Chen,
        Lingyan Ran,
        Peng Wang,
        Yanning Zhang.
        % <-this % stops a space
\thanks{Yinghui Xing, Dexuan Kong, Shizhou Zhang, Geng Chen, Lingyan Ran, Peng Wang and Yanning Zhang are with the School of Computer Science, Northwestern Polytechnical University, Xi’an, China. Yinghui Xing is also with the Research \& Development Institute of Northwestern Polytechnical University in Shenzhen.
}% <-this % stops a space
\thanks{*The first two authors equally contributed to this work.{\dag} correspondence author} 
%\thanks{Manuscript received April 19, 2021; revised August 16, 2021.}
}

\maketitle

%%%%%%%%% ABSTRACT
\begin{abstract}
Camouflaged object detection (COD), aiming to segment camouflaged objects which exhibit similar patterns with the background,
is a challenging task. Most existing works are dedicated to establishing specialized modules
to identify camouflaged objects with complete and fine details, while the boundary can not be well located for the lack of object-related semantics.
In this paper, we propose a novel ``pre-train, adapt and detect" paradigm to detect camouflaged objects.
By introducing a large pre-trained model, abundant knowledge learned from massive multi-modal data
can be directly transferred to COD.
A lightweight parallel adapter is inserted to adjust the features suitable for the downstream COD task. 
Extensive experiments on four challenging benchmark datasets demonstrate that our method outperforms existing state-of-the-art COD models by large margins.
Moreover, we design a multi-task learning scheme for tuning the adapter to exploit the shareable knowledge across different semantic classes. 
Comprehensive experimental results showed that the generalization ability of our model can be substantially improved with multi-task adapter initialization on source tasks and multi-task adaptation on target tasks.
Code will be released.

%Moreover, we designed a multi-task experimental setting to further improve the generalization ability of the proposed model. Through the multi-task learning mechanism, our proposed model not only obtains a further improvement but also avoids interference between different datasets.
\end{abstract}


%-------------------------------------------------------------------------
%%%%%%%%% BODY TEXT
\section{Introduction}

\IEEEPARstart{W}{ild} animals have developed rich camouflage ability, which helps them protect themselves from their predators by blending in with their surroundings~\cite{COD10K/SINet,MGL,ZoomNet}. Camouflaged object detection (COD), typically defined as a binary segmentation task, is more difficult than traditional salient object detection or segmentation, due to the fact that camouflage strategy works by deceiving the
visual perceptual system of the observer, found by sensory ecologists~\cite{stevens2009animal}. 

\begin{figure}[ht!]
    \centering
    \begin{center}
       \includegraphics[width=0.8\linewidth, height=0.55\linewidth]{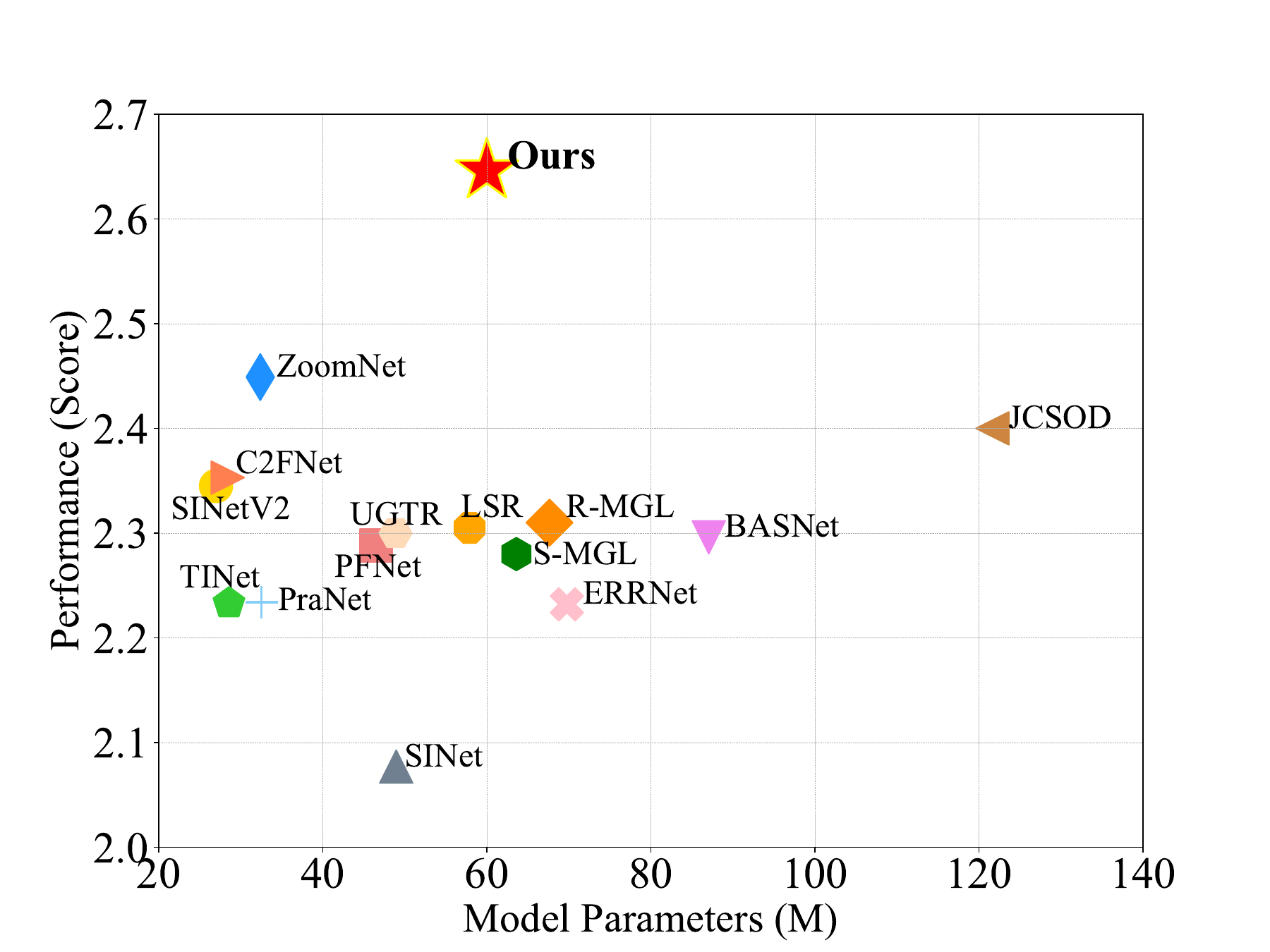}
    \end{center}
    \caption{The scatter relationship between the performance (Score) and parameters of competitors and our model on COD10K-Test~\cite{COD10K/SINet}. $($\textbf{Score} = $S_{\alpha}$ + $E_{\phi}$ + $F^w_\beta$ - M$)$. %Our method achieves a remarkable performance milestone under four widely-used evaluation metrics.
    }
    \label{fig:ParamCmp}
\end{figure}
%-------------------------------------------------------------------------

In recent years, COD has attracted increasing research interest from the computer vision community. 
Early works used low-level hand-crafted features, such as color, edge or texture to detect camouflaged objects~\cite{singh2013survey,huerta2007improving,pan2011study,hou2011detection}, 
%These features are forced to be as discriminative as possible to detect camouflag, 
whose performances are limited for the lack of feature discrimination. 
Le~\etal~\cite{CAMO/le2019anabranch} provided %proposed an anabranch network, 
%which leveraged the strength of both image classification and semantic segmentation tasks to discriminate camouflaged objects, and also provided a 
%together with 
a Camouflaged Object (CAMO) dataset to facilitate the application of deep neural networks to COD task. 
After that, many deep learning based COD methods were proposed. 
Existing research efforts either devoted to devising elaborate modules~\cite{ZoomNet,sun2022boundary} for accurate extraction of object structure or utilized auxiliary tasks
to enhance the discriminative ability of the
main segmentation stream for COD~\cite{zhu2022can, vandenhende2021multi}.
%to simulate the camouflaged object discovery ability of humans~\cite{zhu2022can}.

The major difficulty in COD is how to accurately distinguish the subtle differences between
the target object and the background in the image~\cite{jia2022segment}. However, 
camouflaged objects have a large variety of object appearances, like object size and shape, which further aggravate the accurate detection of boundaries. 
We believe that it can be well addressed by learning context-aware and object-related semantic knowledge~\cite{chen2022camouflaged}.

%Human brain has a much robust vision system, because of its inherent neural architectures with large scale neurons,
With the emergence of large-scale pre-trained models~\cite{MAE,ViT}, many researchers have developed to learn high-quality visual representations by a ``\textit{Pre-training and Fine-tuning}" paradigm~\cite{CoOp,rao2022denseclip}. 
Benefiting from the pre-training, big foundation models with strong generalization ability can be learned with large-scale training data in supervised or self-supervised ways.
Then they can efficiently be adapted to many downstream tasks with lightweight feature adapters~\cite{CLIP-Adapter} or prompts~\cite{CoOp}. 
Like a human that has been seeing and reading countless image samples, the large and deep models can learn and memorize rich general semantic knowledge~\cite{zhu2022prompt}, which we think is favorable to acquire subtle boundaries in COD through the learning of context-aware and object-related semantics.
%can be an imitator to ``intelligent" visual system. 
%With large scale neurons inherently, human vision system once learned can quickly adapt to new perception tasks such as COD.
%Motivated by the camouflage detection mechanism of human vision system, 

In this paper, we propose to solve the COD task from a new perspective via pre-training a large-scale foundation model, then adapting it to COD task with a parameter efficient adapter module.
Specifically, we detect the camouflaged objects within a ``\textit{Pre-train, adapt and detect}” paradigm. 
Large-scale multi-modal data is used to \textit{pre-train} the foundation model, ViT in our implementation, to learn rich meta knowledge. 
Then a lightweight adapter is appended to ViT to \textit{adapt} the pre-trained model to downstream tasks. %To extract multi-scale feature, %for better detection results, 
%the adapter introduces hierarchical feature interaction modules including feature injector and extractor. 
After obtaining a finer feature map, a COD head is used to precisely \textit{detect} the pixel-wise camouflaged object.

Furthermore, it is observed that camouflaged objects have diverse semantic classes, varying from insects, birds, mammals to artificial camouflaged objects. 
Though the classes are diverse, they may have some similar camouflaged patterns, \ie there may be some shareable knowledge among different classes.
It is worth noting that shareable knowledge across different classes may be beneficial to the generalization of the proposed model. Therefore, we further propose to learn the adapter via a multi-task learning paradigm.
%accurate detection of camouflaged objects. 
%To explore the shareable knowledge across different classes, 
%Concretely, multi-task initialization is conducted on multiple source tasks (classes) and multi-task adaptation on target tasks (classes).

We conduct thorough experiments on four widely used benchmark datasets. The proposed %``pre-train, Adapt and Detect'' 
framework outperforms 16 state-of-the-art methods by large margins with less or comparable model parameters, as can be seen from Fig.~\ref{fig:ParamCmp}.
Additionally, to verify the effectiveness of the multi-task adapter tuning, we partitioned the datasets according to the semantic classes of the camouflaged objects. 
Shareable knowledge across tasks can be learned by multi-task adapter initialization on source tasks and multi-task adaptation on target tasks to improve the generalization ability of our method.

The main contributions can be summarized as follows,
\begin{itemize}
    \item We propose to detect camouflaged objects with a novel ``pre-train, adapt and detect" framework. Benefiting from pre-training, the proposed method achieves superior performance by only tuning a small number of parameters without elaborate design. To the best of our knowledge, it is the first COD method based on a large-scale pre-trained foundation model.
    %to solve COD task by ``pre-train, Adapt and Detect''.
    \item We further propose to learn the adapter in a multi-task learning scheme, including multi-task adapter initialization and multitask adapter adaptation. Experiments on zero-shot task transferability, multi-task adaption, and cross-task generalization demonstrate the efficacy of each component.
    \item Our method sets a new record on four widely used benchmark datasets and provides new evaluation protocols to explore multi-task learning on COD tasks.
    \item The proposed method shows strong generalization ability by achieving superior performances on five Salient Object Detection (SOD) benchmark datasets as well.  
\end{itemize}
The rest of the paper is organized as follows, Section~\ref{sec_RW} reviews the relevant works of our paper. Section~\ref{sec_method} elaborates each component of the proposed method. In Section~\ref{sec_expr}, we conduct thorough experiments on four widely-used benchmark datasets to verify the effectiveness of our method. And we draw the conclusion of the paper in Section~\ref{sec_conclusion}

%-------------------------------------------------------------------------
\section{Related Work}\label{sec_RW}
\subsection{Camouflaged Object Detection}
%Camouflage is a useful technique for animals to conceal themselves from visual detection by others. Due to its important scientific and practical value, significant efforts have been made to detect/segment the camouflaged objects from natural scenes. 
Detecting/segmenting camouflaged objects is of great potential in many applications. 
Early works mainly rely on hand-crafted features such as color~\cite{huerta2007improving}, 3D convexity~\cite{pan2011study}, and motion~\cite{hou2011detection}, \textit{etc}. 
These methods work well in a few simple cases but often fail in complex cases, such as scenes with multiple or occluded objects. 
Recent works resort to deep learning to recognize camouflaged objects in more challenging scenarios~\cite{COD10K/SINet,chen2022camouflaged,TINet,TARM}. 
Some of them are based on feature fusion to improve multi-scale object detection performance by capturing rich context information and aggregating cross-level features~\cite{COD10K/SINet, chen2022camouflaged}. 
While others take advantage of the rotation invariant and anti-noise ability of texture features to amplify the difference between camouflaged objects and background~\cite{TINet, TARM}. 
Although these methods improve the performance of camouflaged object detection, they still have limitations in the scenes where the camouflaged objects have high similarity with their background. 
In order to obtain accurate boundary and refined structures, Zhai~\etal~\cite{MGL} introduced graph convolution to capture camouflaged regions and used the Edge-Constricted Graph Reasoning module to explicitly merge edge information. 
Qin~\etal developed a boundary-aware segmentation network (BASNet)~\cite{BASNet}, which learned three-level hierarchy representations through a hybrid loss. 
Due to the camouflage strategy essentially deceiving the visual perception system, edge-based detection still has difficulties to achieve excellent performance. 
To further mimic the behavior of predators in nature or human visual psychological patterns, bio-inspired methods have recently emerged, such as PFNet~\cite{PFNet}, MirrorNet~\cite{Mirrornet} and ZoomNet~\cite{ZoomNet}. 
However these methods imitated human visual systems in a simple manner, which limits their performance. 
Different from the above methods, our model has learned more extensive knowledge through various tasks (including multi-modal data), making it more ``intelligent" to confront the deceits of camouflaged objects.

\subsection{Parameter Efficient Tuning}
Recently, large-scale pre-trained models, such as CLIP~\cite{radford2021learning}, BEIT~\cite{Beit} and Vision Transformer~\cite{ViT}, demonstrate great potential in many natural language processing and computer vision tasks. 
To minimize training costs when adapting large pre-trained models for downstream tasks, parameter efficient tuning focuses on updating only a small subset of parameters for the target task~\cite{Akari}. 
Mainstream methods can be roughly divided into two groups: prompt tuning~\cite{Prompts,CoOp,bahng2022exploring,VPT} and adapter tuning~\cite{Adapters,CLIP-Adapter,Convpass,AdaptFormer}. 

%Prompts~\cite{Prompts} are text instructions that are prepended to the input as a hint to inform the pre-trained language model about the task. Zhou~\etal proposed context optimization (CoOp)~\cite{CoOp}, which avoids manual prompt tuning by modeling context words with continuous vectors that are end-to-end learned from data. VPT~\cite{VPT} introduced a small amount of task-specific learnable parameters into the input sequence of each Transformer layer and learned together with a linear head during fine-tuning. Bahng~\etal~\cite{bahng2022exploring} learned a single image perturbation (i.e., “soft prompt”) to adapt large-scale models in vision.

Adapter tuning~\cite{Adapters} aims to adapt pre-trained models to downstream tasks by inserting a learnable lightweight module, \ie adapter while keeping the pre-trained weights frozen. 
Chen~\etal proposed AdaptFormer~\cite{AdaptFormer}, which set two fully connected layers in parallel to the feed-forward network of ViT model to adapt it to downstream visual recognition tasks. 
%Considering the visual inductive bias (\textit{e.g.}, spatial locality and 2D neighborhood structure), Jie~\etal~\cite{Convpass} designed a ``vision-oriented module" for pre-trained ViT to efficiently capture visual information. 
%Convpass, a simple yet effective PETL method which leverages trainable convolutional blocks as bypasses to adapt pre-trained ViT to downstream visual tasks.
To enhance the few-shot capability of vision-language pre-trained model like CLIP, Gao~\etal~\cite{CLIP-Adapter} proposed to append a lightweight two-layer MLP to the pre-trained fixed-weight CLIP model. 
%While greatly improved the performance of few-shot classification, it still required extra training and computational resources. 
%Zhang~\etal~\cite{Tip-adapter} then developed a Training-Free CLIP-Adapter (Tip-Adapter) to acquire well-performed adapter weights without any training, which is efficient and effective.
%and mixed the original zero-shot visual or language embedding with the corresponding finetuning feature via residual connection. predicted the adapted feature residuals for each input image.

%-------------------------------------------------------------------------
\begin{figure*}[ht!]
    \centering
    \resizebox{\linewidth}{!}{
    \belowrulesep=0pt
    \aboverulesep=0pt
    \includegraphics{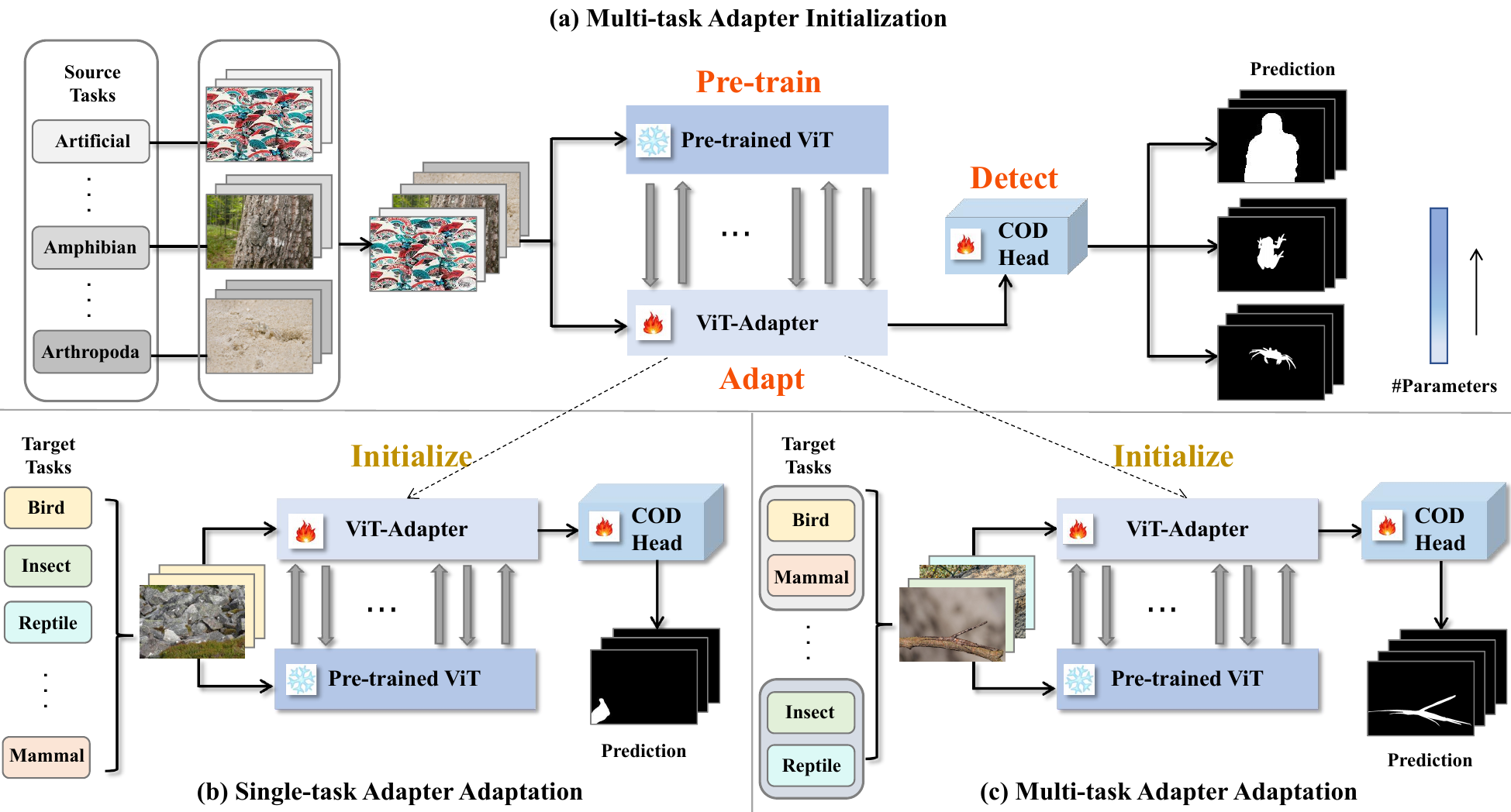}
    }
    \caption{Overall framework. (a) The architecture of the proposed ``Pre-train, Adapt and Detect'' paradigm and multi-task adapter initialization on source tasks. (b) Multi-task adaptation on single target task. (c) Multi-task adaptation on grouped target tasks.}
    \label{fig:Framework}
\end{figure*}
%-------------------------------------------------------------------------
\subsection{Multi-Task Learning}
Learning multi-tasks jointly has been demonstrated as a useful scheme to enable the shared representation or task-specific model to gain more generalized cross-task knowledge~\cite{MTL,vandenhende2021multi}. 
%Multi-Task Learning (MTL)~\cite{MTL,vandenhende2021multi}, a learning paradigm in which multiple tasks are simultaneously learned, aims use the knowledge contained in a task to help the learning of others, whereby improving the generalization performance of all tasks. 
Some auxiliary tasks, such as image classification~\cite{CAMO/le2019anabranch}, salient object detection~\cite{JCSOD}, edge extraction~\cite{MGL}, localization and ranking~\cite{lv2023towards}, have been introduced into the multitask learning paradigm to improve the performance of camouflaged object detection. 
Le~\etal~\cite{CAMO/le2019anabranch} designed a classification branch to predict the probability of containing camouflaged objects in an image. Li~\etal~\cite{JCSOD} presented an adversarial learning framework to conduct multi-task learning on the joint datasets of salient object detection and camouflage object detection. % to enhance the detection ability of them.
%by using the contradictory information between the tasks.
Zhai~\etal~\cite{MGL} decoupled an image into two task-specific feature maps to respectively locate the target and capture boundary details. 
Moreover, Lv~\etal~\cite{lv2023towards} built their network in a triple-task learning framework to simultaneously localize, segment, and rank the camouflaged objects.
Different from these methods that introduced auxiliary functional tasks to excavate extra cues from the shared features, 
we devoted to dividing the COD task into finer sub-tasks according to the semantic labels of the foreground objects, aiming to learn a generalized adapter through cross-task shared knowledge.
%we try to construct subtasks within the COD framework by partitioning the COD datasets into several fine-grained subsets. 
%These methods aim to excavate extra cues from the shared features of other tasks to enhance the feature representation for COD, and this type of multi-task training approaches lack the flexibility of adding or removing source tasks even when some of the tasks cause negative interference.

%This type of multi-task learning can not be well adapted to large-scale pre-trained models.

%Several recent work introduced multi-task fine-tuning into pre-trained models to construct parameter-efficient fine-tuning. Rabeeh~\etal~\cite{Rabeeh} proposed a parameter-efficient method for multitask fine-tuning based on hypernetworks and adapter layers. Akari~\etal~\cite{Akari} introduced a new multi-task, parameter-efficient language model tuning method that learns to transfer knowledge across different tasks via a mixture of soft prompts. Pfeiffer~\etal~\cite{Pfeiffer} designed a novel two-stage transfer learning strategy, termed AdapterFusion, which combines the knowledge from multiple source tasks to perform better on a target task.  
%------------------------------------------------------------s-------------
\section{Methodology}\label{sec_method}
In this section,  we first elaborate on the detailed architecture of the proposed method. Then we introduce a multi-task adapter tuning scheme to make the adapter learn shared knowledge among tasks. 
%We propose to solve the COD task with a three-step process, namely Pre-train, Adapt and Detect, which corresponds to a large pre-trained ViT model, a lightweight adapter module and a detection head respectively. 
%Firstly, we elaborate each component of the architecture. 
%Then, the multi-task adapter tuning scheme is illustrated to enable the adapter to learn shared knowledge among tasks. 

\subsection{Overall Architecture}
We propose to solve the COD task with a three-step process, namely Pre-train, Adapt and Detect, which corresponds to a large pre-trained ViT model, a lightweight adapter module and a detection head respectively. 
The overall architecture is shown in Fig.~\ref{fig:Framework} (a). 
%which consists of three main modules. 
%The first module is a large-scale backbone model. 
%As ViT pre-trained with large-scale multi-modal data~\cite{Uni-perceiver}  can make the obtained features include more semantics, we adopt ViT~\cite{ViT} in our implementation.
%thus the obtained features can have richer semantics.
%process input tokens from various data modalities, we adopt a ViT model pre-trained on large scale multi-modal data~\cite{Uni-perceiver}. 
%The second module is a lightweight pre-training-free vision-specific adapter, namely ViT-Adapter~\cite{vit-adapter}. 
%ViT-adapter includes only a small number of trainable parameters compared with the pre-trained ViT model and serves as a learnable multi-scale feature extracor for downstream COD tasks.
%assisting the general backbone in understanding downstream tasks with the guidance of extra information, using a small number of trainable parameters. 
%The third one is a COD head with a dense prediction branch, and we utilize UPerNet~\cite{UperNet} to detect the camouflaged objects from an extremely similar background.

\textbf{Pre-train.}
Recent researchers come to a consensus to adopt pre-trained models as the backbone for downstream tasks rather than learning models from scratch~\cite{han2021pre}. Although there are many large-scale pre-trained models available, we adopt ViT~\cite{ViT} in our implementation as ViT pre-trained with large-scale multi-modal data~\cite{Uni-perceiver} can make the obtained features include more semantics.
During the pre-training phase, a large-scale ViT model is pre-trained with rich multi-modal data~\cite{Uni-perceiver} and serves as a foundation model.
Following~\cite{Uni-perceiver}, ViT backbone is pre-trained with multi-modal data in terms of image, video and text, since transformer layers can indistinctively process patch embeddings, 3D patch embeddings and token embeddings.
%As transformer layers can indistinctively process patch embeddings, 3D patch embeddings and token embeddings, ViT can be pre-trained with multi-modal data in terms of image, video and text.
%The input image is first fed into the patch embedding, where the image is divided into P×P non-overlapping patches. 
Concretely, all the embeddings are projected into $D$-dimensional vector representations and combined with the positional embeddings. Further, they are fed into $L$ transformer encoder layers.
Note that a learnable ``CLS" token embedding is additionally prepended to extract the global feature representation gradually.
%After that, these patches are flattened and projected to D-dimensional embeddings. Here the feature resolution is reduced to 1/P of the original image. Finally, the overall combination of the patch embeddings and the position embeddings are further fed into L transformer encoder layers for attention calculation.

\textbf{Adapt.} 
After obtaining the foundation model, we need to adapt it for the downstream COD tasks. We introduce a lightweight pre-training-free vision-specific adapter, namely ViT-Adapter~\cite{vit-adapter}, which has only a small number of trainable parameters compared with the pre-trained ViT model, to act as a multi-scale feature extractor for COD task.
%and serves as a learnable multi-scale feature extracor for downstream COD tasks.
As can be seen in Fig.~\ref{fig:Framework}, parallel to the pre-trained ViT, 
%For adaptation, a lightweight adapter module named ViT-Adapter, which is parallel to ViT, is used to extract multi-scale features required by COD task.
ViT-Adapter contains a convolution-based spatial prior module to model the local spatial contexts of input images, and two cross-attention-based modules named as injector and extractor, to make interactions between the adapter and the ViT backbone model. 
After $N$ feature interactions, we obtain fine-grained hierarchical features of similar resolutions to ResNet~\cite{ResNet}, which can be used in COD.
Note that ViT-Adapter includes \textit{fewer than 8\%} parameters of the large-scale pre-trained model.
%The ViT-Adapter is designed as a lightweight module parallel to the ViT, with the total number of parameters fewer than 25M. 

\textbf{Detect.}
%Unified Perceptual Parsing Network (UPerNet).
We use UperNet~\cite{UperNet} as our detection head. The pyramidal features obtained from the adapter are taken into UperNet~\cite{UperNet}. In this process, Pyramid Pooling Module (PPM)~\cite{PPM} is applied on the lowest resolution feature to gain effective global prior representations, and Feature Pyramid Network (FPN)~\cite{FPN} is used to fuse high-level semantic information into middle and low levels via a top-down architecture with lateral connections.  
%We input pyramidal features obtained from adapter into the UperNet by utilizing Pyramid Pooling Module (PPM)~\cite{PPM} and the Feature Pyramid Network (FPN)~\cite{FPN}. 
%PPM is applied on the lowest resolution feature to gain effective global prior representations. 
%And FPN is used to fuse high-level semantic information into middle and low levels via a top-down architecture with lateral connections. 
It could be able to jointly infer and discover the rich visual knowledge underneath images by combining low-resolution, semantically strong features with high-resolution, semantically weak features.

\textbf{Training Process.} 
During training, the camouflaged images are fed into the backbone and the adapter simultaneously. We only optimize the parameters of the adapter module and the detection head while the parameters of the original pre-trained model remain frozen so that the power of the ViT foundation model can be efficiently transferred to the downstream COD task with little computational cost. Our entire training process is supervised by the combination of weighted binary cross-entropy loss ($L^w_{BCE}$)~\cite{f3net} and weighted intersection-over-union loss ($L^w_{IOU}$)~\cite{f3net}, which can be formulated as $L = L^w_{BCE} + L^W_{IOU}$, forcing the model to pay more attention to hard pixels.

\subsection{Multitask Adapter Tuning}
Camouflaged objects have diverse semantic classes, varying from insects to mammals, from natural to artificial camouflage. 
To explore whether shareable knowledge across different semantic classes can be learned via a multi-task learning scheme,
%To explore whether a more generalized adapter can be learned via multi-task learning
%To explore the interaction between different classes of camouflaged objects during model training, 
we divided the CAMO and COD10K datasets into nine non-overlapped sub-datasets as nine different sub-tasks according to the object category. 
We found that compared with the independent learning on a single task, the joint learning across all tasks may improve the performance of the model on most tasks, but it performs poorly on some specific tasks possibly due to task interference/negative transfer issues (Table~\ref{tab:booktabs4}). 
In addition, it is extremely time-consuming to train on all tasks simultaneously. 
In order to improve the generalization performance of the model on each task and reduce resource consumption, we introduced a multi-task learning mechanism, which consists of two stages, multitask source adapter initialization and multitask target adapter adaptation. 

\textbf{Multitask Source Adapter Initialization.}
As shown in Fig.~\ref{fig:Framework} (a),  the adaptation modules are trained jointly on all source tasks in this stage.

\textbf{Multitask Target Adapter Adaptation.}
In this stage, we use the learned source adapter to initialize the target adapter. For single-task target adapter adaptation, we directly tune the adaptation modules on each target task separately, as demonstrated in Fig.~\ref{fig:Framework} (b). 
For multi-task target adapter adaptation shown in Fig.~\ref{fig:Framework} (c), we first group several similar tasks together, then perform multitask adapter tuning within the selected groups. 
The grouping strategy is further discussed in Section~\ref{sec_expr_mtl}. 
Finally, we test the model on each individual target task to evaluate the performance. 

%-------------------------------------------------------------------------
\begin{table*}[ht!]
    \caption{Quantitative comparison with 16 SOTA methods for COD on 4 benchmarks.  $\uparrow$ / $\downarrow$ indicates that larger/smaller is better. \textcolor{red}{\textbf{Red}} and \textcolor{blue}{\textbf{blue}} 
represent the \textbf{first} and \textbf{second} best performing algorithms, respectively. }
    \centering
    \resizebox{\linewidth}{!}{
    \belowrulesep=0pt
    \aboverulesep=0pt
    \begin{tabular}{c|*{4}{c}|*{4}{c}|*{4}{c}|*{4}{c}}
    \toprule
        \multirow{2}*{\textbf{Baseline Models}} & \multicolumn{4}{c|}{\textbf{CHAMELEON}~\cite{CHAMELEON}} & \multicolumn{4}{c|}{\textbf{CAMO-Test}~\cite{CAMO/le2019anabranch}} & \multicolumn{4}{c|}{\textbf{COD10K-Test}~\cite{COD10K/SINet}} & \multicolumn{4}{c}{\textbf{NC4K}~\cite{NC4K/LSR}} \\ 
        \cmidrule(lr){2-5}\cmidrule(lr){6-9}\cmidrule(lr){10-13}\cmidrule(lr){14-17}
         ~ & 
         $S_{\alpha}\uparrow$ & $E_{\phi}\uparrow$ & $F^w_{\beta}\uparrow$ & 
         $~M~\downarrow$ & 
         $S_{\alpha}\uparrow$ & $E_{\phi}\uparrow$ & $F^w_{\beta}\uparrow$ & 
         $~M~\downarrow$ &
         $S_{\alpha}\uparrow$ & $E_{\phi}\uparrow$ & $F^w_{\beta}\uparrow$ & 
         $~M~\downarrow$ &
         $S_{\alpha}\uparrow$ & $E_{\phi}\uparrow$ & $F^w_{\beta}\uparrow$ & 
         $~M~\downarrow$\\
         \cmidrule(lr){1-17} 
        SINet~\cite{COD10K/SINet} & 0.869 & 0.891 & 0.740 & 0.044 & 0.751 & 0.771 & 0.606 & 0.100 & 0.771 & 0.806 & 0.551 & 0.051 & 0.808 & 0.871 & 0.723 & 0.058 \\ 
        PraNet~\cite{PraNet} & 0.860 & 0.898 & 0.763 & 0.044 & 0.769 & 0.824 & 0.663 & 0.094 & 0.789 & 0.861 & 0.629 & 0.045 & 0.822 & 0.876 & 0.724 & 0.059 \\
        TINet~\cite{TINet} & 0.874 & 0.916 & 0.783 & 0.038 & 0.781 & 0.847 & 0.678 & 0.087 & 0.793 & 0.848 & 0.635 & 0.043 & 0.829 & 0.879 & 0.734 & 0.055 \\ 
        ERRNet~\cite{ERRNet} & 0.877 & 0.927 & 0.805 & 0.036 & 0.761 & 0.817 & 0.660 & 0.088 & 0.780 & 0.867 & 0.629 & 0.044 & 0.787 & 0.848 & 0.638 & 0.070 \\
        PFNet~\cite{PFNet} & 0.882 & 0.942 & 0.810 & 0.033 & 0.782 & 0.852 & 0.695 & 0.085 & 0.800 & 0.868 & 0.660 & 0.040 & 0.829 & 0.888 & 0.745 & 0.053 \\ 
        UGTR~\cite{UGTR} & 0.888 & 0.918 & 0.796 & 0.031 & 0.785 & 0.859 & 0.686 & 0.086 & 0.818 & 0.850 & 0.667 & 0.035 & 0.839 & 0.874 & 0.747 & 0.052 \\  
        C$^2$FNet~\cite{C2FNet} & 0.888 & 0.935 & 0.828 & 0.032 & 0.796 & 0.854 & 0.719 & 0.080 & 0.813 & 0.890 & 0.686 & 0.036 & 0.838 & 0.897 & 0.762 & 0.049 \\ 
        SINetV2~\cite{SINetV2} & 0.888 & 0.942 & 0.816 & 0.030 & 0.820 & 0.882 & 0.743 & 0.070 & 0.815 & 0.887 & 0.680 & 0.037 & 0.847 & 0.903 & 0.770 & 0.048 \\                 
        S-MGL~\cite{MGL} & 0.892 & 0.921 & 0.803 & 0.032 & 0.772 & 0.850 & 0.664 & 0.089 & 0.811 & 0.851 & 0.655 & 0.037 & 0.829 & 0.863 & 0.731 & 0.055 \\ 
        R-MGL~\cite{MGL} & 0.893 & 0.923 & 0.813 & 0.030 & 0.775 & 0.847 & 0.673 & 0.088 & 0.814 & 0.865 & 0.666 & 0.035 & 0.833 & 0.867 & 0.740 & 0.052 \\ 
        LSR~\cite{NC4K/LSR} & 0.893 & 0.938 & 0.839 & 0.033 & 0.793 & 0.826 & 0.725 & 0.085 & 0.793 & 0.868 & 0.685 & 0.041 & 0.839 & 0.883 & 0.779 & 0.053 \\  
        JCSOD~\cite{JCSOD} & 0.894 & 0.943 & 0.848 & 0.030 & 0.803 & 0.853 & 0.759 & 0.076 & 0.817 & 0.892 & 0.726 & 0.035 & 0.842 & 0.898 & 0.771 & 0.047 \\  
        ZoomNet~\cite{ZoomNet} & 0.902 & 0.958 & 0.845 & 0.023 & 0.820 & 0.892 & 0.752 & 0.066 & 0.838 & 0.911 & 0.729 & 0.029 & 0.853 & 0.912 & 0.784 & 0.043 \\ 
        BASNet~\cite{BASNet} & \color{blue}{\textbf{0.914}} & 0.954 & 0.866 & \color{blue}{\textbf{0.022}} & 0.749 & 0.796 & 0.646 & 0.096 & 0.802 & 0.855 & 0.677 & 0.038 & 0.817 & 0.859 & 0.732 & 0.058 \\
        HitNet~\cite{HitNet} & \color{red}{\textbf{0.922}} & \color{red}{\textbf{0.970}} & \color{red}{\textbf{0.903}} & \color{red}{\textbf{0.018}} & 0.844 & \color{blue}{\textbf{0.902}} & \color{blue}{\textbf{0.801}} & \color{blue}{\textbf{0.057}} & \color{blue}{\textbf{0.868}} & \color{blue}{\textbf{0.932}} & 0.798 & \color{blue}{\textbf{0.024}} & \color{blue}{\textbf{0.870}} & \color{blue}{\textbf{0.921}} & \color{blue}{\textbf{0.825}} & \color{blue}{\textbf{0.039}} \\ 
        SAM-Adapter~\cite{sam-adapter} & 0.896 & 0.919 & 0.824 & 0.033 & \color{blue}{\textbf{0.847}} & 0.873 & 0.765 & 0.070 & \color{red}{\textbf{0.883}} & 0.918 & \color{blue}{\textbf{0.801}} & 0.025 & - & - & - & - \\
        \cmidrule(lr){1-17}
        \textbf{Ours} & 0.909 & \color{blue}{\textbf{0.959}} & \color{blue}{\textbf{0.891}} & \color{red}{\textbf{0.018}} & \color{red}{\textbf{0.888}} & \color{red}{\textbf{0.942}} & \color{red}{\textbf{0.868}} & \color{red}{\textbf{0.036}} & \color{red}{\textbf{0.883}} & \color{red}{\textbf{0.943}} & \color{red}{\textbf{0.836}} & \color{red}{\textbf{0.016}} & \color{red}{\textbf{0.896}} & \color{red}{\textbf{0.945}} & \color{red}{\textbf{0.874}} & \color{red}{\textbf{0.024}} \\ 
    \bottomrule
    \end{tabular}
    }
    \label{tab:booktabs1}
\end{table*}
%-------------------------------------------------------------------------
\begin{figure*}[ht!]
    \centering
    \resizebox{\linewidth}{!}{
    \belowrulesep=0pt
    \aboverulesep=0pt
    \includegraphics{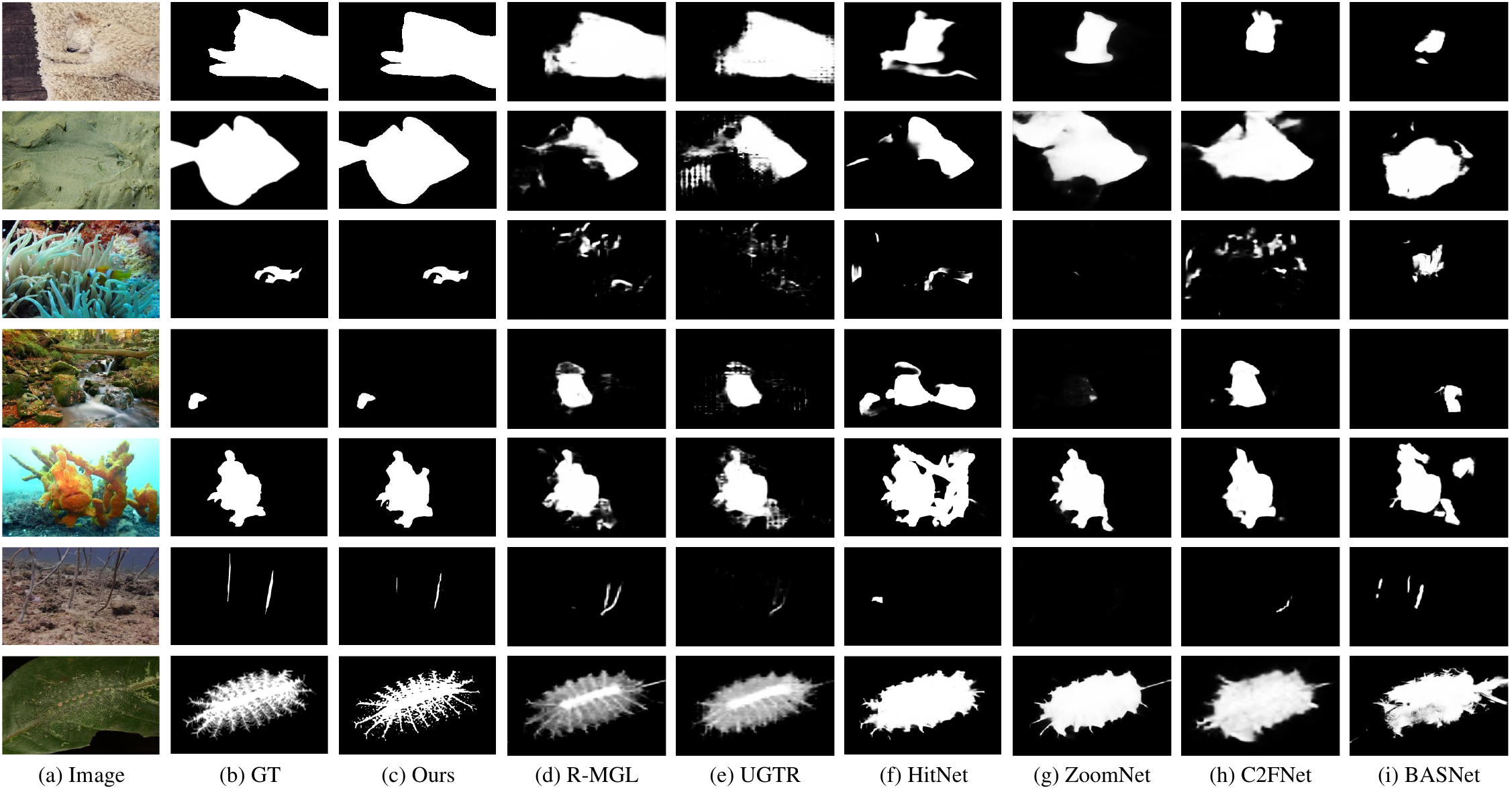}
    }
    \caption{Qualitative comparison of our method with other state-of-the-art COD methods. 
    Our algorithm is capable of tackling challenging cases (\textit{e.g.}, low contrast, occluded, small objects, confusing objects, multiple objects, and complex topological structures).}
    \label{fig:results}
\end{figure*}
%-------------------------------------------------------------------------
\section{Experiments}\label{sec_expr}

We first conduct experiments following the conventional protocols for a fair comparison with current state-of-the-art methods in Section~\ref{expr_ccod}. 
To explore whether shareable knowledge across different semantic classes can be learned via multi-task learning scheme, we then conduct thorough experiments within the multitask framework in Section~\ref{sec_expr_mtl}.

\subsection{Experiments Compared with Other Methods}\label{expr_ccod}
We evaluate our proposed method on both the  Camouflaged object detection (COD) and Salient object detection (SOD) tasks to show the effectiveness and generalization ability of our method in this section.

\textbf{Datasets.}
For camouflaged object detection, our experiments are based on four widely-used COD datasets: 
\textbf{CHAMELEON}~\cite{CHAMELEON}, \textbf{CAMO}~\cite{CAMO/le2019anabranch}, \textbf{COD10K}~\cite{COD10K/SINet} and \textbf{NC4K}~\cite{NC4K/LSR}.
% (1) \textbf{CHAMELEON}~\cite{CHAMELEON} collects 76 images with manually annotated object level ground-truths (GTs). 
% The images were collected from the Internet via the Google search engine using “camouflaged animal” as a keyword. 
% (2) \textbf{CAMO}~\cite{CAMO/le2019anabranch} has 1,250 images (1,000 for training, 250 for testing), covering eight categories, and includes two types of camouflaged objects, namely, natural and artificial camouflage.
% (3) \textbf{COD10K}~\cite{COD10K/SINet} is the largest COD dataset till now, consisting of COD10K-Train (3,040 images) and COD10K-Test (2,026 images). 
% The images are downloaded from multiple free photography websites, covering 5 super-classes and 69 sub-classes. 
% (4) \textbf{NC4K}~\cite{NC4K/LSR}. As the largest testing dataset, NC4K includes 4,121 samples, which are used to evaluate the generalization ability of models. 
Following previous studies~\cite{SINetV2, NC4K/LSR, ZoomNet}, we train our model on the training sets of CAMO and COD10K, and evaluate the detection performance on the whole CHAMELEON and NC4K datasets, together with the test sets of CAMO and COD10K.
% For SOD task, Our experiments are based on five popular SOD datasets:
% (1) \textbf{ECSSD}~\cite{ECSSD} is made up of 1,000 images with meaningful semantics. 
% (2) \textbf{PASCAL-S}~\cite{PASCAL-S} was built from a dataset originally used for semantic segmentation, and it consists of 850 challenging images.
% (3) \textbf{DUTS}~\cite{DUTS} is a relatively large salient object detection dataset with two subsets. The 10,553 images in DUT-TR are used for training, and the 5,019 images
% in DUT-TE are employed for testing.
% (4) \textbf{HKU-IS}~\cite{HKU-IS} includes 4,447 images, which contain multiple foreground objects.
% (5) \textbf{DUT-OMRON}~\cite{OMRON} consists of 5,168 images with at least one object. These objects are usually structurally complex. 
% We follow ~\cite{f3net, ICON} to use DUTS-TR as the training dataset and others as testing datasets.

For salient object detection, we follow~\cite{f3net, ICON} to train our model on the \textbf{DUTS-TR}~\cite{DUTS} dataset, and evaluate on other five testing datasets, including \textbf{ECSSD}~\cite{ECSSD}, \textbf{PASCAL-S}~\cite{PASCAL-S}, \textbf{DUTS-TE}~\cite{DUTS}, \textbf{HKU-IS}~\cite{HKU-IS} and \textbf{DUT-OMRON}~\cite{OMRON}. All datasets are human-labeled with pixel-wise ground-truth for quantitative evaluations.

\textbf{Evaluation Metrics.}
Following existing works~\cite{COD10K/SINet, CAMO/le2019anabranch}, we take four commonly used metrics for evaluation: Structure measure ($S_{\alpha}$)~\cite{Structure-Measure}, Mean enhanced-alignment measure ($E_{\phi}$)~\cite{E-Measure}, weighted F-measure ($F^w_{\beta}$)~\cite{F-Measure}, and mean absolute error ($M$)~\cite{MAE} for both the COD and SOD tasks.

\textbf{Implementation Details.}
In the training phase, we use Vision Transformer~\cite{ViT} as the foundation model and UperNet~\cite{UperNet} as the COD head. 
The Vision Transformer is pre-trained with large-scale multi-modal data as in Uni-Perceiver~\cite{Uni-perceiver} and kept frozen once pre-trained. 
The parameters of adapter and the COD head are both randomly initialized. We employ an AdamW~\cite{AdamW} optimizer with initial learning rate of $6 \times 10^{-5}$ and a weight decay of 0.05. 
They are trained 200 epochs with a batch size of 2. For testing, the images are resized to 512 $\times $512 to input into the model, and the outputs are resized back to the original size.

\textbf{Competitors.}
For camouflaged object detection, we compare our method with 16 recent state-of-the-art methods: SINet~\cite{COD10K/SINet}, PraNet~\cite{PraNet}, TINet~\cite{TINet}, ERRNet~\cite{ERRNet}, PFNet~\cite{PFNet}, UGTR~\cite{UGTR}, C$^2$FNet~\cite{C2FNet}, SINetV2~\cite{SINetV2}, S-MGL~\cite{MGL}, R-MGL~\cite{MGL}, LSR~\cite{NC4K/LSR}, JCSOD~\cite{JCSOD}, ZoomNet~\cite{ZoomNet}, BASNet~\cite{BASNet}, HitNet~\cite{HitNet} and SAM-Adapter~\cite{sam-adapter}. 

For salient object detection, we compare our method with 12 representative methods, including BMPM~\cite{BMPM}, RAS~\cite{RAS}, PiCA-R~\cite{picanet}, DGRL~\cite{DGRL}, CPD-R~\cite{CPD-R}, PoolNet~\cite{PoolNet}, SIBA~\cite{SIBA}, EGNet~\cite{EGNet}, F3Net~\cite{f3net}, ICON~\cite{ICON}, TINet~\cite{TINet} and JCSOD~\cite{JCSOD}.
For fair comparison, all results are either provided in the published paper or reproduced by an open-source model re-trained on the same training set with recommended settings.

\textbf{Quantitative Evaluation.}
The quantitative results for COD task are illustrated in Table~\ref{tab:booktabs1}. As can be seen that our method achieves competitive performance on CHAMELEON dataset and outperforms all other models on other three datasets under four evaluation metrics with large margins, despite only a few parameters are fine-tuned.

%\paragraph{Performance on CAMO.}
For the CAMO test set, our method significantly improves $S_{\alpha}$ by 5.2\%, $E_{\phi}$ by 4.4\%, $F^w_{\beta}$ by 8.4\% and lowers the MAE error by 36.8\%, compared with second-best model HitNet~\cite{HitNet}, which sets a new performance milestone.

%\paragraph{Performance on COD10K.}
For the COD10K test set, our method is consistently better than other competitors. Specifically, compared with the second-best model HitNet~\cite{HitNet}, our model increases $S_{\alpha}$, $E_{\phi}$, and $F^w_{\beta}$ by 1.7\%, 1.2\%, and 4.8\% respectively, and also lowers the MAE error by 33.3\%.

%\paragraph{Performance on NC4K.}
We also evaluate the generalization ability of all models on NC4K dataset. It can be observed from the comparisons in Table~\ref{tab:booktabs1} that our method sets a remarkable record to increase $S_{\alpha}$, $E_{\phi}$, and $F^w_{\beta}$ by 3.0\%, 2.6\%, and 5.9\% respectively, and decreases the MAE error by 38.5\% than the second-best model HitNet~\cite{HitNet}.

Notably, compared to a concurrent work SAM-Adapter~\cite{sam-adapter}, which employs the large pre-trained SAM~\cite{sam} as the foundation model, our method still exhibits obvious superiority.

Table~\ref{tab:booktabs7} reports the quantitative results on five widely used SOD benchmark datasets. It can also be seen that our model performs favorably against the existing methods in terms of nearly all evaluation metrics. This
demonstrates the strong capability and effectiveness of our network to deal with both the COD and SOD tasks.
%-------------------------------------------------------------------------
\begin{table*}[ht!]
    \caption{Quantitative comparison with 12 SOTA methods for SOD on 5 benchmarks.  $\uparrow$ / $\downarrow$ indicates that larger/smaller is better. \textcolor{red}{\textbf{Red}} and \textcolor{blue}{\textbf{blue}} 
represent the \textbf{first} and \textbf{second} best performing algorithms, respectively. }
    \centering
    \resizebox{\linewidth}{!}{
    \belowrulesep=0pt
    \aboverulesep=0pt
    \begin{tabular}{c|*{4}{c}|*{4}{c}|*{4}{c}|*{4}{c}|*{4}{c}}
    \toprule
        \multirow{2}*{\textbf{Baseline Models}} & \multicolumn{4}{c|}{\textbf{ECSSD}~\cite{ECSSD}} & \multicolumn{4}{c|}{\textbf{PASCAL-S}~\cite{PASCAL-S}} & \multicolumn{4}{c|}{\textbf{DUTS-TE}~\cite{DUTS}} & \multicolumn{4}{c|}{\textbf{HKU-IS}~\cite{HKU-IS}} & \multicolumn{4}{c}{\textbf{DUT-OMRON}~\cite{OMRON}} \\ 
        \cmidrule(lr){2-5}\cmidrule(lr){6-9}\cmidrule(lr){10-13}\cmidrule(lr){14-17}
        \cmidrule(lr){18-21}
         ~ & 
         $S_{\alpha}\uparrow$ & $E_{\phi}\uparrow$ & $F^w_{\beta}\uparrow$ & 
         $~M~\downarrow$ & 
         $S_{\alpha}\uparrow$ & $E_{\phi}\uparrow$ & $F^w_{\beta}\uparrow$ & 
         $~M~\downarrow$ &
         $S_{\alpha}\uparrow$ & $E_{\phi}\uparrow$ & $F^w_{\beta}\uparrow$ & 
         $~M~\downarrow$ &
         $S_{\alpha}\uparrow$ & $E_{\phi}\uparrow$ & $F^w_{\beta}\uparrow$ & 
         $~M~\downarrow$ &
         $S_{\alpha}\uparrow$ & $E_{\phi}\uparrow$ & $F^w_{\beta}\uparrow$ & 
         $~M~\downarrow$\\
         \cmidrule(lr){1-21} 
        
        BMPM~\cite{BMPM} & 0.911 & 0.914 & 0.871 & 0.045 & 0.843 & 0.841 & 0.778 & 0.074 & 0.862 & 0.860 & 0.761 & 0.049 & 0.907 & 0.937 & 0.859 & 0.039 & 0.809 & 0.837 & 0.681 & 0.064 \\
        RAS~\cite{RAS} & 0.893 & 0.914 & 0.857 & 0.056 & 0.792 & 0.829 & 0.731 & 0.101 & 0.839 & 0.861 & 0.740 & 0.059 & 0.887 & 0.929 & 0.843 & 0.045 & 0.814 & 0.846 & 0.695 & 0.062 \\
        PiCA-R~\cite{picanet} & 0.917 & 0.913 & 0.867 & 0.046 & 0.848 & 0.832 & 0.772 & 0.074 & 0.869 & 0.862 & 0.754 & 0.043 & 0.904 & 0.936 & 0.840 & 0.043 & 0.832 & 0.841 & 0.695 & 0.065 \\
        DGRL~\cite{DGRL} & 0.906 & 0.917 & 0.903 & 0.043 & 0.834 & 0.836 & 0.807 & 0.074 & 0.846 & 0.863 & 0.764 & 0.051 & 0.896 & 0.941 & 0.881 & 0.037 & 0.810 & 0.843 & 0.709 & 0.063 \\
        CPD-R~\cite{CPD-R} & 0.918 & 0.925 & 0.898 & 0.037 & 0.842 & 0.849 & 0.794 & 0.070 & 0.869 & 0.886 & 0.795 & 0.043 & 0.905 & 0.944 & 0.875 & 0.034 & 0.825 & 0.866 & 0.719 & 0.056 \\
        PoolNet~\cite{PoolNet} & 0.926 & 0.925 & 0.904 & 0.035 & 0.858 & 0.852 & 0.809 & 0.064 & 0.886 & 0.896 & 0.817 & 0.037 & 0.919 & 0.953 & 0.888 & 0.030 & 0.831 & 0.868 & 0.725 & 0.054 \\
        SIBA~\cite{SIBA} & 0.924 & 0.928 & 0.908 & 0.035 & 0.845 & 0.852 & 0.802 & 0.069 & 0.879 & 0.892 & 0.811 & 0.040 & 0.913 & 0.950 & 0.886 & 0.032 & 0.832 & 0.860 & 0.736 & 0.059 \\
        EGNet~\cite{EGNet} & 0.925 & 0.927 & 0.903 & 0.037 & 0.846 & 0.848 & 0.795 & 0.073 & 0.887 & 0.891 & 0.815 & 0.039 & 0.918 & 0.950 & 0.887 & 0.031 & 0.841 & 0.867 & 0.738 & \color{blue}{\textbf{0.053}} \\
        F3Net~\cite{f3net} & 0.924 & 0.925 & 0.912 & 0.034 & 0.854 & 0.859 & 0.816 & 0.061 & 0.888 & 0.902 & 0.835 & 0.035 & 0.917 & 0.953 & 0.900 & 0.028 & 0.838 & 0.870 & 0.747 & \color{blue}{\textbf{0.053}} \\
        ICON~\cite{ICON} & 0.931 & 0.924 & 0.920 & 0.031 & \color{blue}{\textbf{0.864}} & 0.853 & \color{blue}{\textbf{0.830}} & \color{blue}{\textbf{0.059}} & 0.892 & 0.900 & 0.839 & 0.037 & 0.925 & 0.956 & 0.908 & 0.027 & 0.845 & 0.866 & 0.762 & 0.058 \\
        TINet~\cite{TINet} & 0.926 & 0.953 & 0.914 & 0.033 & 0.861 & \color{blue}{\textbf{0.901}} & 0.829 & 0.062 & 0.891 & 0.925 & 0.842 & 0.035 & 0.922 & \color{blue}{\textbf{0.960}} & 0.906 & 0.027 & 0.842 & 0.876 & 0.754 & \color{red}{\textbf{0.051}} \\
        JCSOD~\cite{JCSOD} & \color{blue}{\textbf{0.933}} & \color{blue}{\textbf{0.960}} & \color{blue}{\textbf{0.935}} & \color{blue}{\textbf{0.030}} & - & - & - & - & \color{blue}{\textbf{0.899}} & \color{blue}{\textbf{0.937}} & \color{blue}{\textbf{0.866}} & \color{blue}{\textbf{0.032}} & \color{blue}{\textbf{0.931}} & 0.867 & \color{blue}{\textbf{0.924}} & \color{blue}{\textbf{0.026}} & \color{blue}{\textbf{0.850}} & \color{red}{\textbf{0.884}} & \color{blue}{\textbf{0.782}} & \color{red}{\textbf{0.051}} \\
        \cmidrule(lr){1-21}
        \textbf{Ours} & \color{red}{\textbf{0.943}} & \color{red}{\textbf{0.971}} & \color{red}{\textbf{0.949}} & \color{red}{\textbf{0.019}} & \color{red}{\textbf{0.885}} & \color{red}{\textbf{0.934}} & \color{red}{\textbf{0.872}} & \color{red}{\textbf{0.044}} & \color{red}{\textbf{0.914}} & \color{red}{\textbf{0.952}} & \color{red}{\textbf{0.899}} & \color{red}{\textbf{0.025}} & \color{red}{\textbf{0.934}} & \color{red}{\textbf{0.972}} & \color{red}{\textbf{0.938}} & \color{red}{\textbf{0.018}} & \color{red}{\textbf{0.851}} & \color{blue}{\textbf{0.881}} & \color{red}{\textbf{0.794}} & 0.054 \\  
    \bottomrule
    \end{tabular}
    }
    \label{tab:booktabs7}
\end{table*}
%-------------------------------------------------------------------------
\begin{figure*}[ht!]
    \centering
    \resizebox{\linewidth}{!}{
    \belowrulesep=0pt
    \aboverulesep=0pt
    \includegraphics{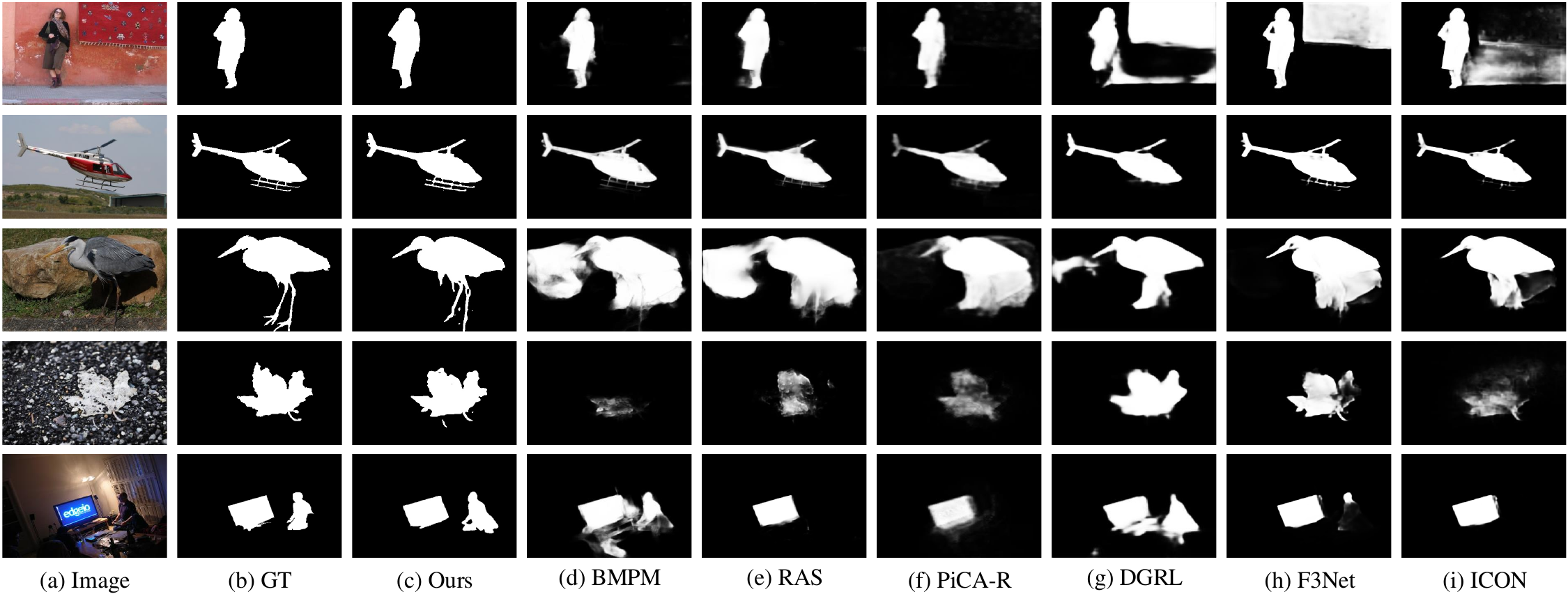}
    }
    \caption{Qualitative comparison of our method with other state-of-the-art SOD methods. Our algorithm is capable of tackling challenging cases (\textit{e.g.}, background interference, fine construction, clutter, unclear boundary and dim light).}
    \label{fig:results_SOD}
\end{figure*}
%-------------------------------------------------------------------------

\textbf{Qualitative Evaluation.}
Fig.~\ref{fig:results} shows the qualitative comparison of several other recent models and ours on COD task. 
As can be seen, our method ($3^{rd}$ column) is able to handle different types of challenging camouflaged cases. 
For objects with low contrast ($1^{st}$ and $2^{nd}$ row), other models can vaguely identify a small part of them, while our method can completely and clearly detect the objects. 
For targets occluded by fine objects ($3^{rd}$ row), we can present more complete prediction than other methods. 
For small objects ($4^{th}$ row), extremely confusing objects ($5^{th}$ row) and multiple objects with low contrast ($6^{th}$ row), our method provides accurate camouflaged object predictions, while 
%at the same time other results 
others are interfered more or less,
%by other objects, 
resulting in wrong locations. 
For the examples that have complex topological structures with lots of dense edges or details ($7^{th}$ row), although they are difficult to be detected even by humans,  our method is capable of segmenting clear edges and boundaries while all other methods failed.

Fig.~\ref{fig:results_SOD} shows the qualitative comparison between our approach and
the representative methods on SOD task. 
As can be observed, our method predicts more accurate saliency maps in various challenging cases, \textit{e.g.},
background interference ($1^{st}$ row), fine construction ($2^{nd}$ row), clutter ($3^{rd}$ row), unclear boundary ($4^{th}$ row) and dim light ($5^{th}$ row). 
Besides, our method can detect salient targets integrally and noiselessly. 
The above results demonstrate the accuracy and robustness of the proposed framework.
%-------------------------------------------------------------------------
\begin{table*}[ht!]
    \caption{Comparison of different pre-trained weights in the case of freezing the backbone. $\uparrow$ and $\downarrow$ indicate the higher score the better and the lower the score the better, respectively. L/B/S/T denotes Large/Base/Small/Tiny respectively.}
    \centering
    \resizebox{\linewidth}{!}{
    \belowrulesep=0pt
    \aboverulesep=0pt
    \begin{tabular}{c|c|*{4}{c}|*{4}{c}|*{4}{c}|*{4}{c}}
    \toprule
        \multirow{2}*{\textbf{Method}} & \multirow{2}*{\textbf{Pre\_train}}  & \multicolumn{4}{c|}{\textbf{CHAMELEON}~\cite{CHAMELEON}} & \multicolumn{4}{c|}{\textbf{CAMO-Test}~\cite{CAMO/le2019anabranch}} & \multicolumn{4}{c|}{\textbf{COD10K-Test}~\cite{COD10K/SINet}} & \multicolumn{4}{c}{\textbf{NC4K}~\cite{NC4K/LSR}} \\ 
        \cmidrule(lr){3-6}\cmidrule(lr){7-10}\cmidrule(lr){11-14}\cmidrule(lr){15-18}
         ~ & ~ & 
         $S_{\alpha}\uparrow$ & $E_{\phi}\uparrow$ & $F^w_{\beta}\uparrow$ & 
         $~M~\downarrow$ & 
         $S_{\alpha}\uparrow$ & $E_{\phi}\uparrow$ & $F^w_{\beta}\uparrow$ & 
         $~M~\downarrow$ &
         $S_{\alpha}\uparrow$ & $E_{\phi}\uparrow$ & $F^w_{\beta}\uparrow$ & 
         $~M~\downarrow$ &
         $S_{\alpha}\uparrow$ & $E_{\phi}\uparrow$ & $F^w_{\beta}\uparrow$ & 
         $~M~\downarrow$\\
         \cmidrule(lr){1-18}
        
        HitNet~\cite{HitNet} & PVT-V2~\cite{PVT-V2}  
         & \textbf{0.922} & \textbf{0.970} & \textbf{0.903} & \textbf{0.018} & 0.844 & 0.902 & 0.801 & 0.057 & 0.868 & 0.932 & 0.798 & 0.024 & 0.870 & 0.921 & 0.825 & 0.039 \\ 
         \cmidrule(lr){1-18}
        \multirow{8}*{\textbf{Ours}} & \multirow{1}*{BEiT-L~\cite{Beit}} & 0.895 & 0.951 & 0.869 & 0.022 & 0.866 & 0.922 & 0.841 & 0.047 & 0.866 & 0.921 & 0.810 & 0.018 & 0.884 & 0.931 & 0.858 & 0.026 \\ 
        % ~ & ~ & 364.6M & 0.921 & 0.965 & 0.908 & 0.015 & 0.905 & 0.955 & 0.896 & 0.028 & 0.901 & 0.952 & 0.869 & 0.013 & 0.909 & 0.952 & 0.895 & 0.021 \\ 
        ~ & \multirow{1}*{Uni-Per-L~\cite{Uni-perceiver}} & 0.909 & 0.959 & 0.891 & \textbf{0.018} & \textbf{0.888} & \textbf{0.942} & \textbf{0.868} & \textbf{0.036} & \textbf{0.883} & \textbf{0.943} & \textbf{0.836} & \textbf{0.016} & \textbf{0.896} & \textbf{0.945} & \textbf{0.874} & \textbf{0.024} \\ 
        % ~ & ~ & 363.3M & 0.917 & 0.969 & 0.907 & 0.015 & 0.880 & 0.936 & 0.860 & 0.038 & 0.883 & 0.936 & 0.836 & 0.016 & 0.894 & 0.936 & 0.869 & 0.025 \\ 
        ~ & \multirow{1}*{AugReg-L~\cite{AugReg}} & 0.869 & 0.931 & 0.820 & 0.027 & 0.831 & 0.889 & 0.789 & 0.056 & 0.830 & 0.881 & 0.750 & 0.024 & 0.859 & 0.905 & 0.817 & 0.033 \\ 
        % ~ & ~ & 363.7M & 0.910 & 0.962 & 0.898 & 0.017 & 0.885 & 0.941 & 0.868 & 0.035 & 0.879 & 0.932 & 0.832 & 0.016 & 0.890 & 0.932 & 0.863 & 0.024 \\
        ~ & \multirow{1}*{AugReg-B~\cite{AugReg}} & 0.860 & 0.922 & 0.807 & 0.031 & 0.843 & 0.911 & 0.806 & 0.053 & 0.837 & 0.904 & 0.757 & 0.024 & 0.869 & 0.926 & 0.829 & 0.032 \\ 
        % ~ & ~ & 133.8M & 0.902 & 0.950 & 0.879 & 0.019 & 0.859 & 0.920 & 0.831 & 0.044 & 0.858 & 0.911 & 0.792 & 0.020 & 0.874 & 0.920 & 0.836 & 0.029 \\ 
        ~ & \multirow{1}*{DeiT-B~\cite{Deit}} & 0.858 & 0.915 & 0.812 & 0.032 & 0.823 & 0.886 & 0.783 & 0.057 & 0.819 & 0.871 & 0.731 & 0.026 & 0.850 & 0.901 & 0.804 & 0.036 \\ 
        % ~ & ~ & 133.5M & 0.895 & 0.949 & 0.872 & 0.019 & 0.837 & 0.902 & 0.805 & 0.056 & 0.841 & 0.897 & 0.763 & 0.023 & 0.859 & 0.907 & 0.814 & 0.033 \\ 
        ~ & \multirow{1}*{DeiT-S~\cite{Deit}} & 0.848 & 0.908 & 0.795 & 0.031 & 0.793 & 0.858 & 0.738 & 0.071 & 0.801 & 0.859 & 0.696 & 0.031 & 0.833 & 0.888 & 0.774 & 0.042 \\ 
        % ~ & ~ & 57.5M & 0.885 & 0.938 & 0.851 & 0.022 & 0.817 & 0.884 & 0.776 & 0.061 & 0.825 & 0.883 & 0.735 & 0.026 & 0.850 & 0.898 & 0.798 & 0.037 \\ 
        ~ & \multirow{1}*{AugReg-T~\cite{AugReg}} & 0.809 & 0.893 & 0.730 & 0.042 & 0.751 & 0.816 & 0.673 & 0.085 & 0.776 & 0.847 & 0.654 & 0.036 & 0.817 & 0.875 & 0.751 & 0.048 \\ 
        % ~ & ~ & 35.9M & 0.851 & 0.906 & 0.789 & 0.031 & 0.759 & 0.819 & 0.686 & 0.081 & 0.784 & 0.838 & 0.660 & 0.035 & 0.813 & 0.861 & 0.740 & 0.048 \\ 
        ~ & \multirow{1}*{DeiT-T~\cite{Deit}} & 0.816 & 0.900 & 0.748 & 0.042 & 0.767 & 0.837 & 0.699 & 0.089 & 0.785 & 0.859 & 0.671 & 0.036 & 0.821 & 0.883 & 0.755 & 0.048 \\ 
        % ~ & ~ & 35.9M & 0.859 & 0.916 & 0.804 & 0.027 & 0.757 & 0.821 & 0.684 & 0.081 & 0.776 & 0.843 & 0.647 & 0.039 & 0.805 & 0.859 & 0.730 & 0.052 \\
        \bottomrule
    \end{tabular}
    }
    \label{tab:booktabs2}
\end{table*}

\begin{table*}[ht!]
    \caption{Comparison of different pre-trained weights in the case of unfreezing the backbone. $\uparrow$ and $\downarrow$ indicate the higher score the better and the lower the score the better, respectively. L/B/S/T denotes Large/Base/Small/Tiny respectively.}
    \centering
    \resizebox{\linewidth}{!}{
    \belowrulesep=0pt
    \aboverulesep=0pt
    \begin{tabular}{c|c|*{4}{c}|*{4}{c}|*{4}{c}|*{4}{c}}
    \toprule
        \multirow{2}*{\textbf{Method}} & \multirow{2}*{\textbf{Pre\_train}}  & \multicolumn{4}{c|}{\textbf{CHAMELEON}~\cite{CHAMELEON}} & \multicolumn{4}{c|}{\textbf{CAMO-Test}~\cite{CAMO/le2019anabranch}} & \multicolumn{4}{c|}{\textbf{COD10K-Test}~\cite{COD10K/SINet}} & \multicolumn{4}{c}{\textbf{NC4K}~\cite{NC4K/LSR}} \\ 
        \cmidrule(lr){3-6}\cmidrule(lr){7-10}\cmidrule(lr){11-14}\cmidrule(lr){15-18}
         ~ & ~ & 
         $S_{\alpha}\uparrow$ & $E_{\phi}\uparrow$ & $F^w_{\beta}\uparrow$ & 
         $~M~\downarrow$ & 
         $S_{\alpha}\uparrow$ & $E_{\phi}\uparrow$ & $F^w_{\beta}\uparrow$ & 
         $~M~\downarrow$ &
         $S_{\alpha}\uparrow$ & $E_{\phi}\uparrow$ & $F^w_{\beta}\uparrow$ & 
         $~M~\downarrow$ &
         $S_{\alpha}\uparrow$ & $E_{\phi}\uparrow$ & $F^w_{\beta}\uparrow$ & 
         $~M~\downarrow$\\
         \cmidrule(lr){1-18}
        
        HitNet~\cite{HitNet} & PVT-V2~\cite{PVT-V2}
         & \textbf{0.922} & \textbf{0.970} & 0.903 & 0.018 & 0.844 & 0.902 & 0.801 & 0.057 & 0.868 & 0.932 & 0.798 & 0.024 & 0.870 & 0.921 & 0.825 & 0.039 \\ \cmidrule(lr){1-18}
        \multirow{8}*{\textbf{Ours}} & \multirow{1}*{BEiT-L~\cite{Beit}} & 
        0.921 & 0.965 & \textbf{0.908} & \textbf{0.015} & \textbf{0.905} & \textbf{0.955} & \textbf{0.896} & \textbf{0.028} & \textbf{0.901} & \textbf{0.952} & \textbf{0.869} & \textbf{0.013} & \textbf{0.909} & \textbf{0.952} & \textbf{0.895} & \textbf{0.021} \\ 
        ~ & \multirow{1}*{Uni-Per-L~\cite{Uni-perceiver}} & 
         0.917 & 0.969 & 0.907 & 0.015 & 0.880 & 0.936 & 0.860 & 0.038 & 0.883 & 0.936 & 0.836 & 0.016 & 0.894 & 0.936 & 0.869 & 0.025 \\ 
        ~ & \multirow{1}*{AugReg-L~\cite{AugReg}} & 
        0.910 & 0.962 & 0.898 & 0.017 & 0.885 & 0.941 & 0.868 & 0.035 & 0.879 & 0.932 & 0.832 & 0.016 & 0.890 & 0.932 & 0.863 & 0.024 \\
        ~ & \multirow{1}*{AugReg-B~\cite{AugReg}} & 
        0.902 & 0.950 & 0.879 & 0.019 & 0.859 & 0.920 & 0.831 & 0.044 & 0.858 & 0.911 & 0.792 & 0.020 & 0.874 & 0.920 & 0.836 & 0.029 \\ 
        ~ & \multirow{1}*{DeiT-B~\cite{Deit}} & 
        0.895 & 0.949 & 0.872 & 0.019 & 0.837 & 0.902 & 0.805 & 0.056 & 0.841 & 0.897 & 0.763 & 0.023 & 0.859 & 0.907 & 0.814 & 0.033 \\ 
        ~ & \multirow{1}*{DeiT-S~\cite{Deit}} & 
        0.885 & 0.938 & 0.851 & 0.022 & 0.817 & 0.884 & 0.776 & 0.061 & 0.825 & 0.883 & 0.735 & 0.026 & 0.850 & 0.898 & 0.798 & 0.037 \\ 
        ~ & \multirow{1}*{AugReg-T~\cite{AugReg}} & 
        0.851 & 0.906 & 0.789 & 0.031 & 0.759 & 0.819 & 0.686 & 0.081 & 0.784 & 0.838 & 0.660 & 0.035 & 0.813 & 0.861 & 0.740 & 0.048 \\ 
        ~ & \multirow{1}*{DeiT-T~\cite{Deit}} & 
        0.859 & 0.916 & 0.804 & 0.027 & 0.757 & 0.821 & 0.684 & 0.081 & 0.776 & 0.843 & 0.647 & 0.039 & 0.805 & 0.859 & 0.730 & 0.052 \\
        \bottomrule
    \end{tabular}
    }
    \label{tab:booktabs3}
\end{table*}

\begin{table*}[ht!]
    \caption{Comparison of different detection heads. The backbone is initialized with the Uni-Perceiver-L~\cite{Uni-perceiver} and keep frozen. $\uparrow$ and $\downarrow$ indicate the higher score the better and the lower the score the better, respectively.}
    \centering
    \resizebox{\linewidth}{!}{
    \belowrulesep=0pt
    \aboverulesep=0pt
    \begin{tabular}{c|c|*{4}{c}|*{4}{c}|*{4}{c}|*{4}{c}}
    \toprule
        \multirow{2}*{\textbf{Method}} & \multirow{2}*{\textbf{Detection Head}}  & \multicolumn{4}{c|}{\textbf{CHAMELEON}~\cite{CHAMELEON}} & \multicolumn{4}{c|}{\textbf{CAMO-Test}~\cite{CAMO/le2019anabranch}} & \multicolumn{4}{c|}{\textbf{COD10K-Test}~\cite{COD10K/SINet}} & \multicolumn{4}{c}{\textbf{NC4K}~\cite{NC4K/LSR}} \\ 
        \cmidrule(lr){3-6}\cmidrule(lr){7-10}\cmidrule(lr){11-14}\cmidrule(lr){15-18}
         ~ & ~ & 
         $S_{\alpha}\uparrow$ & $E_{\phi}\uparrow$ & $F^w_{\beta}\uparrow$ & 
         $~M~\downarrow$ & 
         $S_{\alpha}\uparrow$ & $E_{\phi}\uparrow$ & $F^w_{\beta}\uparrow$ & 
         $~M~\downarrow$ &
         $S_{\alpha}\uparrow$ & $E_{\phi}\uparrow$ & $F^w_{\beta}\uparrow$ & 
         $~M~\downarrow$ &
         $S_{\alpha}\uparrow$ & $E_{\phi}\uparrow$ & $F^w_{\beta}\uparrow$ & 
         $~M~\downarrow$\\
         \cmidrule(lr){1-18}
        
        HitNet~\cite{HitNet} & None
         & \textbf{0.922} & 0.970 & \textbf{0.903} & 0.018 & 0.844 & 0.902 & 0.801 & 0.057 & 0.868 & 0.932 & 0.798 & 0.024 & 0.870 & 0.921 & 0.825 & 0.039 \\ \cmidrule(lr){1-18}
        \multirow{8}*{\textbf{Ours}} & 
            \multirow{1}*{FPNHead~\cite{FPN}} & 
        0.915 & \textbf{0.974} & 0.900 & \textbf{0.017} & 0.888 & 0.944 & 0.871 & 0.036 & \textbf{0.889} & 0.951 & \textbf{0.841} & \textbf{0.016} & 0.902 & 0.950 & 0.881 & \textbf{0.023} \\ 
        ~ & \multirow{1}*{ISAHead ~\cite{isanet}} & 
        0.912 & 0.973 & 0.892 & 0.018 & \textbf{0.889} & \textbf{0.949} & \textbf{0.873} & \textbf{0.035} & \textbf{0.889} & \textbf{0.956} & \textbf{0.841} & \textbf{0.016} & \textbf{0.904} & \textbf{0.954} & \textbf{0.883} & \textbf{0.023} \\
        ~ & \multirow{1}*{SETRPUPHead~\cite{setr}} & 
        0.911 & 0.973 & 0.893 & 0.019 & 0.886 & 0.944 & 0.867 & 0.037 & 0.878 & 0.948 & 0.824 & 0.018 & 0.894 & 0.947 & 0.869 & 0.025 \\
        ~ & \multirow{1}*{SETRMLAHead~\cite{setr}} & 
        0.910 & 0.971 & 0.892 & 0.019 & 0.883 & 0.940 & 0.862 & 0.039 & 0.879 & 0.948 & 0.826 & 0.018 & 0.895 & 0.947 & 0.871 & 0.025 \\
        ~ & \multirow{1}*{OCRHead~\cite{ocr}} & 
        0.909 & 0.972 & 0.892 & 0.019 & 0.885 & 0.945 & 0.865 & 0.036 & 0.883 & 0.946 & 0.831 & 0.017 & 0.899 & 0.949 & 0.874 & 0.024 \\ 
        ~ & \multirow{1}*{PSPHead~\cite{psp}} & 
        0.909 & 0.965 & 0.887 & 0.019 & 0.885 & 0.946 & 0.869 & 0.037 & 0.884 & 0.950 & 0.833 & 0.017 & 0.901 & 0.951 & 0.879 & 0.024 \\
        ~ & \multirow{1}*{UperHead~\cite{UperNet}} & 
        0.909 & 0.959 & 0.891 & 0.018 & 0.888 & 0.942 & 0.868 & 0.036 & 0.883 & 0.943 & 0.836 & \textbf{0.016} & 0.896 & 0.945 & 0.874 & 0.024 \\ 
        ~ & \multirow{1}*{SegformerHead~\cite{segformer}} & 
        0.896 & 0.965 & 0.870 & 0.021 & 0.877 & 0.938 & 0.854 & 0.041 & 0.878 & 0.948 & 0.822 & 0.019 & 0.894 & 0.947 & 0.866 & 0.026 \\
        \bottomrule
    \end{tabular}
    }
    \label{tab:booktabs6}
\end{table*}
%----------------------------------------------------------------------------

\textbf{Ablation Study.}
In this experiment, we first study the effect on 4 different foundation models (ViT-Large/Base/Small/Tiny) pre-trained with 4 different ways, namely DeiT~\cite{Deit} which incorporates a data-efficient ViT pre-trained on ImageNet-1K, AugReg~\cite{AugReg} which borrows lots of data augmentation and regularization technique and is trained on ImageNet-22K, BEiT~\cite{Beit} which is pre-trained with a self-supervised way on ImageNet-1K, and Uni-Perceiver~\cite{Uni-perceiver} which is trained with large scale multi-modal data.
In addition, we investigate the effect of adapter-tuning only and full fine-tuning with the frozen and unfrozen backbone parameters. 
The results are provided in Table~\ref{tab:booktabs2} and Table~\ref{tab:booktabs3}. 
We initialize ViT-Tiny/Small/Base with the DeiT~\cite{Deit} released weights, and ViT-Tiny/Base/Large with the ImageNet-22K weights from~\cite{AugReg}. 
We also use the BEiT~\cite{Beit} pre-trained weights and Uni-Perceiver~\cite{Uni-perceiver} to initialize the ViT-L separately.\footnote{The pre-trained model weights with other settings are not released.} 

In the case of frozen backbone, as shown in Table~\ref{tab:booktabs2}, using Uni-Perceiver~\cite{Uni-perceiver} initialization shows the best results that greatly outperform existing SOTA methods. 
While in the case of unfrozen backbone, as shown in Table~\ref{tab:booktabs3}, it achieves astonishing full fine-tuning results when using BEiT~\cite{Beit} initialization, which greatly outperforms the existing counterparts. 
Specifically, our method boosts $S_{\alpha}$ by 7.2\%, $E_{\phi}$ by 5.9\%, $F^w_{\beta}$ by 11.9\% and lowers the MAE error by 50.9\% on CAMO dataset; increases $S_{\alpha}$ by 3.8\%, $E_{\phi}$ by 2.1\%, $F^w_{\beta}$ by 8.9\% and lowers the MAE error by 45.8\% on COD10K dataset; and improves $S_{\alpha}$ by 4.5\%, $E_{\phi}$ by 3.4\%, $F^w_{\beta}$ by 8.5\% and lowers the MAE error by 46.2\% on NC4K dataset. 
We found that fully fine-tuning models significantly outperform the counterpart method when just tuning adapter in most cases. 
However, the latter is much more efficient with less than 8\% parameters.
Notably, in Section~\ref{sec_expr_mtl}, we will try to close the gap between adapter-tuning only and full fine-tuning through a multi-task learning mechanism.

Besides, we also explore the impact of detection head on the proposed framework. 
Specifically, we initialize the large foundation model with the Uni-Perceiver-L released weights~\cite{Uni-perceiver} and keep the components in our framework unchanged except for the detection head. The experimental results for 8 different detection heads are displayed in Table~\ref{tab:booktabs6}. It can be seen that the performance obtained by using different detection heads varies slightly and all surpass existing SOTA methods on most benchmark datasets. 
This shows the robustness and stability of our proposed framework.

\subsection{Multitask Learning Experiments}\label{sec_expr_mtl}
%To explore whether shareable knowledge across different semantic classes can be learned via multi-task learning scheme, we conduct thorough experiments based on multitask learning.
Multi-task learning mechanism is mainly evaluated in the following three problem settings: 
(1) Zero-shot task transferability; 
(2) Multitask adaptation, which shows the effectiveness of multitask adapter initialization and adaption; 
(3) Cross-task generalization, which measures the generalization ability of multitask adapter. 
%Before discussing the results, we provide the details of the experimental setup below.

\textbf{Datasets.}
We divided the entire COD10K datasets and the artificial camouflaged part of CAMO datasets into nine non-overlapped sub-datasets as nine different tasks according to the object category, which includes Amphibian, Arthropoda, Artificial, Bird, Insect, Mammal, Reptile, Underwater1 and Underwater2 (Underwater1 and Underwater2 have obvious differences). 
For the first two settings, we use all nine tasks as source tasks and target tasks. For the cross-task generalization setting, we randomly sample five tasks as source tasks and the remaining four non-overlapped tasks as targets. 

%\textbf{Baselines.}
%(1) The adaptation modules are initialized randomly and independently trained on each target task while keeping the backbone frozen. 
%(2) The adaptation modules are initialized randomly and independently trained on each target task while backbone parameters are also fine-tuned. 
%(3) The adaptation modules are initialized randomly and jointly trained on all tasks while keeping the backbone frozen. 
%(4) The adaptation modules are initialized randomly and jointly trained on all tasks while backbone parameters are also fine-tuned.

\textbf{Implementation Details.}
Throughout the multi-task experiments, we use ViT-Large as the backbone and initialize it with the Uni-Perceiver-L released weights~\cite{Uni-perceiver}. 
We train the adaptation modules on source tasks for 100 epochs and on target tasks for 200 epochs with a batch size of 2. 
%During target adapter tuning, we save a checkpoint every 50 epochs and report results on the checkpoint with the highest validation performance. 
All the input images is resized to 512×512. We use AdamW~\cite{AdamW} optimizer with an initial learning rate of $6 \times 10^{-5}$ and a weight decay of 0.05.
%-------------------------------------------------------------------------
\begin{table*}[ht!]
    \caption{Comparison of adapter learning methods on 9 target tasks. $($\textbf{Score} = $S_{\alpha}$ + $E_{\phi}$ + $F^w_\beta$ - M$)$. %\textbf{Bold} font indicates the best performance.
    }
    \centering
    \resizebox{\linewidth}{!}{
    \belowrulesep=0pt
    \aboverulesep=0pt
    \begin{tabular}{cc|*{9}{c}|c}
    \toprule
        \multicolumn{2}{c|}{\multirow{2}*{\textbf{Method}}} & \multicolumn{9}{c|}{\textbf{Target Dataset}} & \multirow{2}*{\textbf{Avg}} \\ 
        % \Xcline{2-10}{0.1pt}
        \cmidrule(lr){3-11}
        ~ & ~ & Amphibian & Arthropoda & Artificial & Bird & Insect & Mammal & Reptile & Underwater1 & Underwater2 & ~ \\ 
        
        % \midrule 
        % \rowcolor{gray!20}\multicolumn{11}{c}{\it \textbf{Adapter-tuning}}\\
        \midrule
        \multicolumn{1}{c|}{\multirow{4}*{\it \textbf{Adapter-tuning}}}
        & ST & 2.731 & 2.497 & 2.498 & 2.728 & 2.640 & 2.543 & 2.598 & 2.579 & 2.618 & 2.604 \\ 
        \multicolumn{1}{c|}{} & MT & 2.725 & 2.496 & 2.594 & 2.725 & 2.633 & 2.552 & \textbf{2.652} & \textbf{2.609} & 2.657 & 2.627 \\ 
        \multicolumn{1}{c|}{} & MS-ST & 2.742 & 2.506 & 2.558 & 2.735 & 2.640 & 2.558 & 2.621 & 2.584 & 2.639 & 2.620 \\ 
        \multicolumn{1}{c|}{} & MS-MT & \textbf{2.759} & \textbf{2.540} & \textbf{2.596} & \textbf{2.752} & \textbf{2.645} & \textbf{2.597} & 2.647 & 2.586 & \textbf{2.691} & \textbf{2.646} \\ 
        % \midrule 
        % \rowcolor{gray!20}\multicolumn{11}{c}{\it \textbf{Model-tuning}}\\
        \midrule
        \multicolumn{1}{c|}{\multirow{4}*{\it \textbf{Model-tuning}}}
        & ST & 2.753 & 2.574 & 2.643 & 2.744 & 2.681 & 2.595 & 2.645 & 2.598 & 2.636 & 2.652 \\ 
        \multicolumn{1}{c|}{} & MT & 2.709 & 2.545 & 2.636 & 2.753 & 2.670 & \textbf{2.597} & 2.658 & 2.617 & 2.675 & 2.651 \\ 
        \multicolumn{1}{c|}{} & MS-ST & 2.733 & 2.580 & \textbf{2.666} & \textbf{2.755} & 2.660 & 2.591 & \textbf{2.668} & 2.623 & \textbf{2.697} & 2.664 \\ 
        \multicolumn{1}{c|}{} & MS-MT & \textbf{2.769} & \textbf{2.594} & 2.638 & \textbf{2.755} & \textbf{2.686} & 2.589 & 2.654 & \textbf{2.630} & 2.690 & \textbf{2.667} \\ 
    \bottomrule
    \end{tabular}
    }
    \label{tab:booktabs4}
\end{table*}
%---------------------------------------------------------------------------
\begin{figure}[ht!]
    \centering
    \centering
    \begin{center}
       \includegraphics[width=0.85\linewidth]{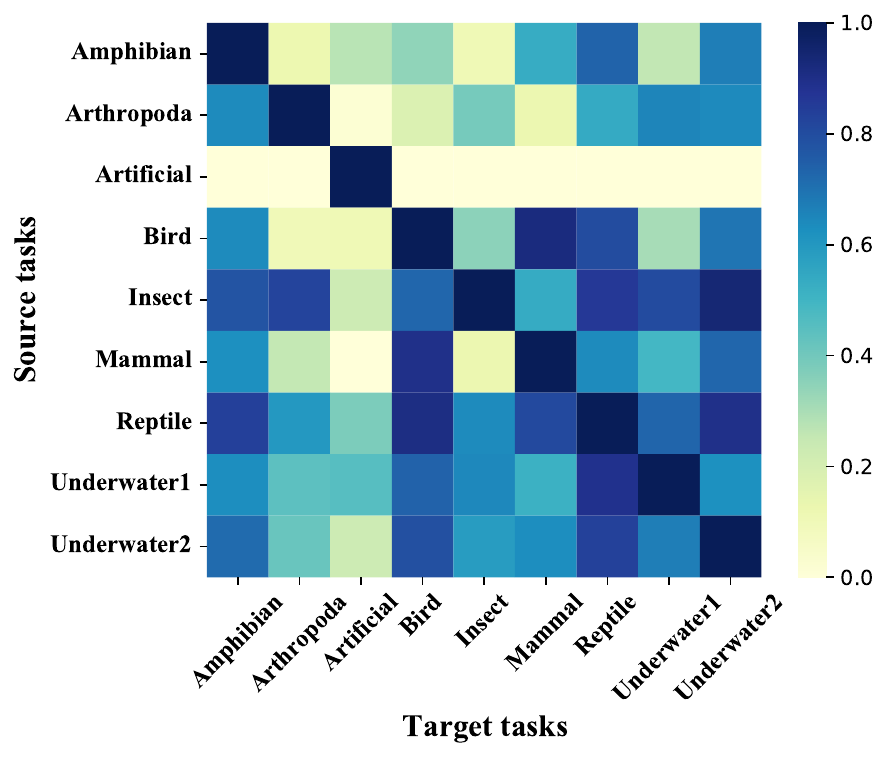}
    \end{center}
    \caption{A heatmap of our task transferability results. Each cell shows the relative performance on the target tasks of the transferred adapter from the associated source task (row) to the associated target task (column).}
    \label{fig:heatmap}
\end{figure}
%-------------------------------------------------------------------------

\textbf{Zero-shot Task Transferability.}
Which tasks should be learned together to help share information among tasks? 
To investigate this, we conduct a large-scale empirical study with 9 tasks in 81 combinations. 
We perform 200 epochs tuning on each source task to initialize
%Then the checkpoint with the highest source task validation performance is selected to initialize 
adapter for different target tasks followed by zero-shot adaptation. 
We normalize the scores to the range of [0, 1] by dividing the transfer performance with the best one on that task and presented the results in Fig.~\ref{fig:heatmap}. 
It demonstrates that tasks with similar characteristics have better transferability, such as Bird, Mammal and Reptile. 
To determine the appropriate grouping strategy, we select the top-3 transferability tasks for each target task. 
Then we jointly train such groups
%of three tasks
in multitask adaptation stage 
%and select the best checkpoint based on the validation performance 
to test each task respectively.

\begin{table*}[ht!]
    \caption{The results of five group experiments about cross-task generalization. Note that there is no overlap between source and target tasks, where "-" indicates source tasks. $($\textbf{Score} = $S_{\alpha}$ + $E_{\phi}$ + $F^w_\beta$ - M$)$.}
    \centering
    \resizebox{\linewidth}{!}{
    \belowrulesep=0pt
    \aboverulesep=0pt
    \begin{tabular}{c|c|c*{9}{c}}
    \toprule
        \multirow{2}*{\textbf{Method}} & \multirow{2}*{\textbf{Group}} & \multicolumn{9}{c}{\textbf{Target Dataset}} \\ 
        % \Xcline{2-10}{0.1pt}
        \cmidrule(lr){3-11}
        ~ & ~ & Amphibian & Arthropoda & Artificial & Bird & Insect & Mammal & Reptile & Underwater1 & Underwater2 \\ 
        \midrule 
        ST & None & 2.731 & 2.497 & 2.498 & 2.728 & 2.640 & 2.543 & 2.598 & 2.579 & 2.618 \\
        \midrule 
        \multirow{5}*{\textbf{MS\_ST}} & 1 & - & - & - & - & \textbf{2.653} & - & \textbf{2.619} & 2.585 & \textbf{2.627} \\         
        ~ & 2 & 2.749 & - & - & 2.737 & - & \textbf{2.561} & 2.599 & - & - \\        
        ~ & 3 & 2.744 & - & \textbf{2.531} & - & - & - & 2.602 & \textbf{2.586} & - \\       
        ~ & 4 & \textbf{2.750} & \textbf{2.530} & 2.522 & \textbf{2.738} & - & - & - & - & - \\        
        ~ & 5 & - & 2.518 & 2.530 & - & \textbf{2.653} & 2.550 & - & - & - \\   
    \bottomrule
    \end{tabular}
    }
    \label{tab:booktabs5}
\end{table*}

\textbf{Multitask Adaptation.}
We present results for all 9 subtasks in Table~\ref{tab:booktabs4}. 
Single-Task (ST) refers to the results of independently trained adapter/model for each task. 
Multi-Task (MT) refers to the results of jointly trained adapter/model on all tasks. 
Multi-Source-Task-Single-Target-Task (MS-ST) shows the results of single-task target adapter/model adaptation using the initialization adapter on multi-source tasks. 
Multi-Source-Task-Multi-Target-Task (MS-MT) shows the results of multi-task target adapter/model adaptation using the initialization adapter on multi-source tasks. 

By comparing the performances of different fine-tuning approaches in the case of \textit{\textbf{adapter-tuning}} in Table~\ref{tab:booktabs4}, we summarized four important findings:
(1) Comparing ST and MT, MT boosts the averaged performance by 0.9\%, which demonstrates that there are advantages in sharing knowledge cross tasks through multi-task learning. However, we also find that there are considerable performance drops for Amphibian and Insect using MT over ST, which indicates that learning a shareable adapter jointly can not guarantee the best results for all tasks.
(2) By comparing ST and MS-ST, together with MT and MS-MT, we observe that multi-source-task initialization %both MS-ST and MS-MT are improved, 
improves the detection accuracy, verifying the efficacy of multitask adapter initialization. Specifically, MS-ST outperforms ST on averaged score by 0.6\% and increases the performance for 8/9 tasks. MS-MT boosts 0.7\% on averaged score than MT, and improves the performance for 7/9 tasks, which shows that leveraging information from other tasks has favorable effects.
(3) In contrast to MS-ST, MS-MT obtains better results on all tasks and improves the results 1.0\% on average. It clearly demonstrates that multitask adapter adaptation variants exhibit better transferability than single target task adapter adaptation counterparts. 
(4) Furthermore, we can see that MS-MT under adapter-tuning almost achieves the performance of ST under full model-tuning, despite only a small number of parameters of the entire model are updated. It proves that our multi-task learning mechanism can reduce the gap with full model-tuning and even exceeds full model-tuning by a large margin on some tasks such as Reptile and Underwater2. %while retaining all the computational benefits.

\textbf{Cross-Task Generalization.}
We conduct five groups of experiments in which the source tasks and target tasks are not overlapped. 
We perform multi-task learning on the source tasks to learn a shared adapter. 
Then the shared adapter is used as the adapter initialization for single-task adaptation on each target task. 
The results shown in Table~\ref{tab:booktabs5} indicate that multi-task adapter initialization mostly outperform the baseline adapter learning counterparts by a significant margin. 
It concludes that using the adapter trained on the source tasks to initialize the adapter of the target task can help the target task better learn the knowledge transferred from the source tasks, and also help to improve the generalization ability of the model.
%-------------------------------------------------------------------------

\section{Conclusion}\label{sec_conclusion}
In this paper, a novel ``pre-train, adapt and detect" paradigm is proposed to detect camouflaged objects.
A large scale foundation model is pre-trained with massive multi-modal data, with the assumption that abundant knowledge can be learned through the pre-training step and can be efficiently transferred to benefit the downstream COD task.
A lightweight adapter with much fewer tunnable parameters is devised to adjust the features of the foundation model to suit for dense prediction COD task.
We conduct extensive experiments on four challenging COD benchmark datasets and five SOD benchmark datasets, and the results demonstrate that our method outperforms existing state-of-the-art COD and SOD models by a large margin under four widely-used evaluation metrics. 
Moreover, we designed a multi-task learning scheme to explore whether the adapter can learn shareable knowledge across tasks.
Comprehensive experimental results verified that multi-task adapter initialization on source tasks and adaptation on target tasks can further improve the generalization ability of the proposed model. 

%Through the multi-task learning mechanism, our proposed model not only obtains a further improvement, but also avoids the interference between different datasets.

\bibliographystyle{IEEEtran}
\bibliography{egbib}

\vfill

\end{document}